\documentclass{article}

\usepackage{arxiv}

\usepackage[utf8]{inputenc} 
\usepackage[T1]{fontenc}    
\usepackage{hyperref}       
\usepackage{url}            
\usepackage{booktabs}       
\usepackage{amsfonts}       
\usepackage{nicefrac}       
\usepackage{microtype}      
\usepackage{lipsum}
\usepackage{graphicx}
\usepackage{amsmath}
\usepackage{algorithm}
\usepackage{algorithmic}
\usepackage{siunitx}
\usepackage{acronym}

\graphicspath{ {./images/} }

\title{Hessian Aware Quantization of Spiking Neural Networks}

\author{
  Hin Wai Lui \\
  Department of Computer Science\\
  University of California, Irvine\\
  Irvine, CA 92697 \\
  \texttt{hwlui@uci.edu} \\
   \And
  Emre Neftci \\
  Department of Computer Science\\
  University of California, Irvine\\
  Irvine, CA 92697 \\
  \texttt{eneftci@uci.edu} \\
}

\begin{document}
\acrodef{IR}[IR]{Intrinsic Rewards and Motivation}
\acrodef{PPO}[PPO]{Proximal Policy Optimization}
\acrodef{RL}[RL]{Reinforcement Learning}
\acrodef{AC}[AC]{Arrenhius \& Current}
\acrodef{AD}[AD]{Automatic Differentiation}
\acrodef{AER}[AER]{Address Event Representation}
\acrodef{AEX}[AEX]{AER EXtension board}
\acrodef{AMDA}[AMDA]{``AER Motherboard with D/A converters''}
\acrodef{ANN}[ANN]{Artificial Neural Network}
\acrodef{API}[API]{Application Programming Interface}
\acrodef{BP}[BP]{Back-Propagation}
\acrodef{BPTT}[BPTT]{Back-Propagation-Through-Time}
\acrodef{BM}[BM]{Boltzmann Machine}
\acrodef{CAVIAR}[CAVIAR]{Convolution AER Vision Architecture for Real-Time}
\acrodef{CCN}[CCN]{Cooperative and Competitive Network}
\acrodef{CD}[CD]{Contrastive Divergence}
\acrodef{CG}[CG]{Computational Graph}
\acrodef{CMOS}[CMOS]{Complementary Metal--Oxide--Semiconductor}
\acrodef{CNN}[CNN]{Convolutional Neural Network}
\acrodef{COTS}[COTS]{Commercial Off-The-Shelf}
\acrodef{CPU}[CPU]{Central Processing Unit}
\acrodef{CV}[CV]{Coefficient of Variation}
\acrodef{CTC}[CTC]{connectionist temporal classification}
\acrodef{DAC}[DAC]{Digital--to--Analog}
\acrodef{DBN}[DBN]{Deep Belief Network}
\acrodef{DCLL}[DECOLLE]{Deep Continuous Local Learning}
\acrodef{DFA}[DFA]{Deterministic Finite Automaton}
\acrodef{DFA}[DFA]{Deterministic Finite Automaton}
\acrodef{divmod3}[DIVMOD3]{divisibility of a number by 3}
\acrodef{DPE}[DPE]{Dynamic Parameter Estimation}
\acrodef{DNN}[DNN]{Deep Neural Network}
\acrodef{DPI}[DPI]{Differential-Pair Integrator}
\acrodef{DSP}[DSP]{Digital Signal Processor}
\acrodef{DVS}[DVS]{Dynamic Vision Sensor}
\acrodef{EDVAC}[EDVAC]{Electronic Discrete Variable Automatic Computer}
\acrodef{EIF}[EI\&F]{Exponential Integrate \& Fire}
\acrodef{EIN}[EIN]{Excitatory--Inhibitory Network}
\acrodef{EPSC}[EPSC]{Excitatory Post-Synaptic Current}
\acrodef{EPSP}[EPSP]{Excitatory Post--Synaptic Potential}
\acrodef{eRBP}[eRBP]{Event-Driven Random Back-Propagation}
\acrodef{FPGA}[FPGA]{Field Programmable Gate Array}
\acrodef{FSM}[FSM]{Finite State Machine}
\acrodef{GPU}[GPU]{Graphical Processing Unit}
\acrodef{HAL}[HAL]{Hardware Abstraction Layer}
\acrodef{HH}[H\&H]{Hodgkin \& Huxley}
\acrodef{HMM}[HMM]{Hidden Markov Model}
\acrodef{HW}[HW]{Hardware}
\acrodef{hWTA}[hWTA]{Hard Winner--Take--All}
\acrodef{IF2DWTA}[IF2DWTA]{Integrate \& Fire 2--Dimensional WTA}
\acrodef{IF}[I\&F]{Integrate \& Fire}
\acrodef{IFSLWTA}[IFSLWTA]{Integrate \& Fire Stop Learning WTA}
\acrodef{INCF}[INCF]{International Neuroinformatics Coordinating Facility}
\acrodef{INRC}[INRC]{Intel Neuromorphic Research Community}
\acrodef{INI}[INI]{Institute of Neuroinformatics}
\acrodef{IO}[IO]{Input-Output}
\acrodef{IoT}[IoT]{internet of things}
\acrodef{IPSC}[IPSC]{Inhibitory Post-Synaptic Current}
\acrodef{ISI}[ISI]{Inter--Spike Interval}
\acrodef{JFLAP}[JFLAP]{Java - Formal Languages and Automata Package}
\acrodef{LIF}[LIF]{Leaky Integrate-and-Fire}
\acrodef{LSM}[LSM]{Liquid State Machine}
\acrodef{LTD}[LTD]{Long-Term Depression}
\acrodef{LTI}[LTI]{Linear Time-Invariant}
\acrodef{LTP}[LTP]{Long-Term Potentiation}
\acrodef{LTU}[LTU]{Linear Threshold Unit}
\acrodef{LSTM}[LSTM]{long short-term memory}
\acrodef{MCMC}{Markov Chain Monte Carlo}
\acrodef{MSE}{Mean-Squared Error}
\acrodef{NHML}[NHML]{Neuromorphic Hardware Mark-up Language}
\acrodef{NMDA}[NMDA]{NMDA}
\acrodef{NME}[NE]{Neuromorphic Engineering}
\acrodef{PCB}[PCB]{Printed Circuit Board}
\acrodef{PRC}[PRC]{Phase Response Curve}
\acrodef{PSC}[PSC]{Post-Synaptic Current}
\acrodef{PSP}[PSP]{Post--Synaptic Potential}
\acrodef{RI}[KL]{Kullback-Leibler}
\acrodef{RRAM}[RRAM]{Resistive Random-Access Memory}
\acrodef{RBM}[RBM]{Restricted Boltzmann Machine}
\acrodef{RTRL}[RTRL]{Real-Time Recurrent Learning}
\acrodef{ROC}[ROC]{Receiver Operator Characteristic}
\acrodef{RSA}[RSA]{Representational Similarity Analysis}
\acrodef{RDA}[RDA]{Representational Dissimilarity Analysis}
\acrodef{RDM}[RDA]{Representational Dissimilarity Matrix}
\acrodef{RNN}[RNN]{Recurrent Neural Network}
\acrodef{SAC}[SAC]{Selective Attention Chip}
\acrodef{SCD}[SCD]{Spike-Based Contrastive Divergence}
\acrodef{SCX}[SCX]{Silicon CorteX}
\acrodef{SG}[SG]{Surrogate Gradient}
\acrodef{SGD}[SGD]{Surrogate Gradient Descent}
\acrodef{SRM}[SRM]{Spike Response Model}
\acrodef{SNN}[SNN]{Spiking Neural Network}
\acrodef{STDP}[STDP]{Spike Time Dependent Plasticity}
\acrodef{SW}[SW]{Software}
\acrodef{sWTA}[SWTA]{Soft Winner--Take--All}
\acrodef{TPU}[TPU]{Tensorflow Processing Unit}
\acrodef{VHDL}[VHDL]{VHSIC Hardware Description Language}
\acrodef{VLSI}[VLSI]{Very  Large  Scale  Integration}
\acrodef{WTA}[WTA]{Winner--Take--All}
\acrodef{XML}[XML]{eXtensible Mark-up Language}

\maketitle
\begin{abstract}
To achieve the low latency, high throughput, and energy efficiency benefits of \acp{SNN}, reducing the memory and compute requirements when running on a neuromorphic hardware is an important step. 
Neuromorphic architecture allows massively parallel computation with variable and local bit-precisions.
However, how different bit-precisions should be allocated to different layers or connections of the network is not trivial.
In this work, we demonstrate how a layer-wise Hessian trace analysis can measure the sensitivity of the loss to any perturbation of the layer's weights, and this can be used to guide the allocation of a layer-specific bit-precision when quantizing an \ac{SNN}.
In addition, current gradient based methods of \ac{SNN} training use a complex neuron model with multiple state variables, which is not ideal for compute and memory efficiency.
To address this challenge, we present a simplified neuron model that reduces the number of state variables by 4-fold while still being compatible with gradient based training. 
We find that the impact on model accuracy when using a layer-wise bit-precision correlated well with that layer's Hessian trace. 
The accuracy of the optimal quantized network only dropped by 0.3\%, yet the network size was reduced by 58\%. This reduces memory usage and allows fixed-point arithmetic with simpler digital circuits to be used, increasing the overall throughput and energy efficiency.
\end{abstract}


\section{Introduction} \label{introduction}
Spiking Neural Networks (SNNs) promise low latency, high throughput, energy efficient, and event-driven information processing when paired with appropriate neuromorphic hardware~\cite{Pfeiffer2018}. 
Unlike \acp{DNN} with dense activity patterns and dense matrix multiplications, \acp{SNN} process and communicate information with sparse discrete spikes and are naturally equipped with internal memory states (\emph{i.e.} they are recurrent neural networks).


Dedicated neuromorphic hardware can realize the benefits of spike based computation of SNNs. 
For this work, we focus on digital implementations of such hardware as they are more widespread today.
Neuromorphic hardware allows massively parallel computation of neuron dynamics, with memory local to computation, which prevents the Von Neumann bottleneck~\cite{Indiveri2011}. 
To make efficient use of energy and silicon area, digital implementations of neuromorphic hardware often only provide integer or low-precision arithmetic. For example, the TrueNorth chip~\cite{Esser_etal16_convnetw} uses 2-bit trinary weights and performs classification tasks with frame rates exceeding 1,200 frames/s using only $25$mW.
The Loihi chip~\cite{Davies_etal18_loihneur} supports synaptic plasticity states (\emph{e.g.} weights) with bit-precisions ranging between 1 and 9 bits.
Furthermore, the bit-precision of the weights on the Loihi is connection-specific, and this capability is only possible with the parallel computation and local memory architecture of Neuromorphic hardware.
Traditional Von Neumann or Single Instruction Multiple Data (SIMD) processors such as CPUs or GPUs cannot efficiently provide connection-specific bit-precision with dense weight matrices, due to the entailing instruction overhead.
On the other hand, there is evidence to suggest that the brain adapts such variable weight bit-precision strategy, through synapses with variable sizes~\cite{Bartol2015}.
The locality and adaptability of computation and memory resources are what inspired the field of neuromorphic engineering at the first place.

Existing hardware implementations and investigations on neural network quantization~\cite{Schaefer2020, Esser2019, Esser_etal16_convnetw} argued that integer arithmetic could significantly reduce the memory usage while only marginally increasing the error, provided that the quantization is accounted for during training.
\cite{Esser2019} showed that fine-tuning a quantized \ac{DNN} that was fully trained with full precision weights could achieve full precision accuracy, when a learnt quantizer step size was used. 
\cite{Schaefer2020} achieved near full precision accuracy on a quantized \ac{SNN}  with the same fine-tuning approach, using a uniform bit-precision with 10-12 bits assigned for neuron state variables, and 6 bits for the weights.
However, these approaches cannot take full advantage of the connection-specific bit-precision capability provided by neuromorphic hardware.
Recently, a Hessian aware approach showed good results on mixed precision quantization of \acp{DNN}~\cite{Dong2019}. 
It allowed different layers of the network to have a different bit-precision, guided by the Hessian trace of that layer.
The layer-wise Hessian trace provides a measure of the sensitivity of the loss to any perturbation of the layer's parameters. 
Therefore, a layer with a low Hessian trace could afford to have a lower bit-precision than one with a high Hessian trace. 
With this approach, each layer within the network could use an optimally low bit-precision with only small impact on the loss.

In this work, we explore the use of Hessian aware quantization on \acp{SNN}, and adapt the Hutchinson method for fast estimations of Hessian trace on \acp{SNN}.
We also present a simplified version of a \ac{LIF} neuron model that could further compress the memory usage and speed up computation. 
We introduce this neuron model in section~\ref{dynamcis}, the method of quantization in section~\ref{quantization}, the method of computing the Hessian trace of an SNN in section~\ref{hessian}, the experiments and results in section~\ref{results}, and conclusion and future work in section~\ref{conculsion}.
\section{Simplified Neuron Model} \label{dynamcis}
The standard model of an \ac{LIF} neuron~\cite{Gerstner2014} has the following form:
\begin{equation} 
\begin{split}
\label{eqn:lif}
\rho_i[n] & = \alpha U_i[n-1]S_i[n-1], \\
U_i[n] & = \alpha U_i[n-1] + \sum_{j} W_{ij} S_j[n-1] - \rho_i[n], \\
S_i[n] & = \Theta (U_i[n] - U_{thres}), \\
\end{split}
\end{equation}
where $U_i[n]$ is the internal membrane potential of the post-synaptic neuron at time step $n$, $S_j$ is the pre-synaptic spike, $S_i$ is the post-synaptic spike, $W_{ij}$ is the synaptic weight between pre and post-synaptic neuron, $\alpha$ is the exponential decay constant, $U_{thresh}$ is the reset threshold, and $\Theta$ is the step function. 
The post-synaptic neuron receives pre-synaptic spikes and integrates them to its membrane potential $U_i$.
When $U_i$ is above the threshold $U_{thresh}$, it generates a spike $S_i$ at that time step.
For an LIF neuron, there is a leakage current that results in $U_i$ decaying by the factor $\alpha$ between each time step.
In this model, the neuron resets its potential to $0$ after a spike is generated. 
Here we represent this reset behavior with the reset variable $\rho_i$, which is the decayed membrane potential just before the spike generation.
It is subtracted from $U_i$ after the spike generation to reset $U_i$ back to $0$.  

In this article, we use Deep Continuous Local Learning to train the SNN. 
DECOLLE enables the scalable training of \acp{SNN} locally with very long time sequences.
DECOLLE uses an approximation of forward-mode autodiff in which $\frac{\partial U_i}{\partial W_{ij}}$ are traces of past pre-synaptic spiking activities~\cite{Zenke2021} and gradients are computed using layer-wise local classifiers. 
Implementing DECOLLE~\cite{Kaiser2020} with Equation (\ref{eqn:lif}) and automatic differentiation is not straightforward.
Instead, DECOLLE uses a discrete-time Spike Response Model formulation of Equation (\ref{eqn:lif}) \cite{Gerstner_etal14_neurdyna} which consists of 5 state variables, the eligibility trace of the current based synapse $Q_j$, the membrane potential trace of the pre-synaptic neuron $P_j$, the reset dynamical variable of the post-synaptic neuron $R_i$, the membrane potential $U_i$, and the spike train $S_i$.
While the DECOLLE neuron model offers some level of biological realism by including synaptic dynamics, it requires more memory and computation than compared to Equation (\ref{eqn:lif}).
Therefore, we modify it to the following:
\begin{equation}
\label{eqn:srm}
\begin{split}
P_j[n] & = \alpha P_j[n-1] + S_j[n-1], \\
R_i[n] & = \alpha R_i[n-1] + \alpha U_i[n-1]S_i[n-1], \\
U_i[n] & = \sum_{j}W_{ij}P_j[n] - R_i[n], \\
S_i[n] & = \Theta(U_i[n] - U_{thres}). \\
\end{split}
\end{equation}
Equation (\ref{eqn:srm}) is simplified from the DECOLLE neuron model with the state variable $Q_i$ and the factor $(1-\alpha)$ removed. It is also equivalent to (\ref{eqn:lif}). We can prove this by subtracting the state variables of (\ref{eqn:srm}) between two subsequent time steps:
\begin{equation}
\label{eqn:P-diff}
P_j[n] - \alpha P_j[n-1] =  S_j[n-1],
\end{equation}
\begin{equation}
\label{eqn:Q-diff}
R_i[n] - \alpha R_i[n-1] = \alpha U_i[n-1]S_i[n-1] = \rho_i[n],
\end{equation}
\begin{equation}
\label{eqn:U-diff}
U_i[n] - \alpha U_i[n-1] = \sum_{j}W_{ij}(P_j[n] - \alpha P_j[n-1]) + (R_i[n] - \alpha R_i[n-1]).
\end{equation}
Substituting (\ref{eqn:P-diff}) and (\ref{eqn:Q-diff}) into (\ref{eqn:U-diff}):
\begin{equation}
\label{eqn:prove}
U_i[n] = \alpha U_i[n-1] + \sum_{j}W_{ij} S_j   [n-1] - \rho_i[n].
\end{equation}
Equation (\ref{eqn:prove}) has the same form as $U_i$ of (\ref{eqn:lif}).
The difference is that (\ref{eqn:srm}) allows us to compute $\frac{\partial U_i}{\partial W_{ij}}$ as the pre-synaptic traces without gradient backpropagation, which makes it compatible with DECOLLE:
\begin{equation}
\label{eqn:du_dt}
\frac{\partial U_i}{\partial W_{ij}} = P_j[n],
\end{equation}
Hence the loss gradient with respect to the weight $\frac{\partial \mathcal{L}}{\partial W_{ij}}$ is:
\begin{equation}
\label{eqn:dL_dt}
\begin{split}
\frac{\partial \mathcal{L}}{\partial W_{ij}} & = \frac{\partial \mathcal{L}}{\partial S_i} \frac{\partial S_i}{\partial U_i} \frac{\partial U_i}{\partial W_{ij}} \\
& = E_i[n] \sigma'(U_i[n])P_j[n],
\end{split}
\end{equation}
where $E_i[n]$ is the propagated loss gradient to the neuron, and $\sigma'$ is the surrogate gradient function. 
This neuron model allows us to use (\ref{eqn:srm}) during training, and (\ref{eqn:lif}) during inference to gain the benefits of reduced memory usage and computation. Furthermore, since the reset mechanism of (\ref{eqn:lif}) can be implemented by simply setting $U_i=0$, the state variable $\rho_i$ is not actually needed during inference, further simplifying the computation. Overall, only one state variable, the internal membrane potential $U_i$, is needed during inference.

\section{State and weight quantization} \label{quantization}
Low precision arithmetic can bring a number of benefits to \ac{SNN} running on a neuromorphic hardware. Synaptic weight quantization can reduce the memory usage and storage space required, allowing for a higher throughput given the same memory bandwidth. If both the neuron state variables and synaptic weights are quantized to low precision fixed-point representation, simpler digital circuits with less silicon can be used. 
This allows more neurons to be simulated in parallel within the same silicon and power budget, increasing the overall throughput and energy efficiency.

\begin{figure}[htbp]
 	\centering
	\includegraphics[width=0.8\linewidth]{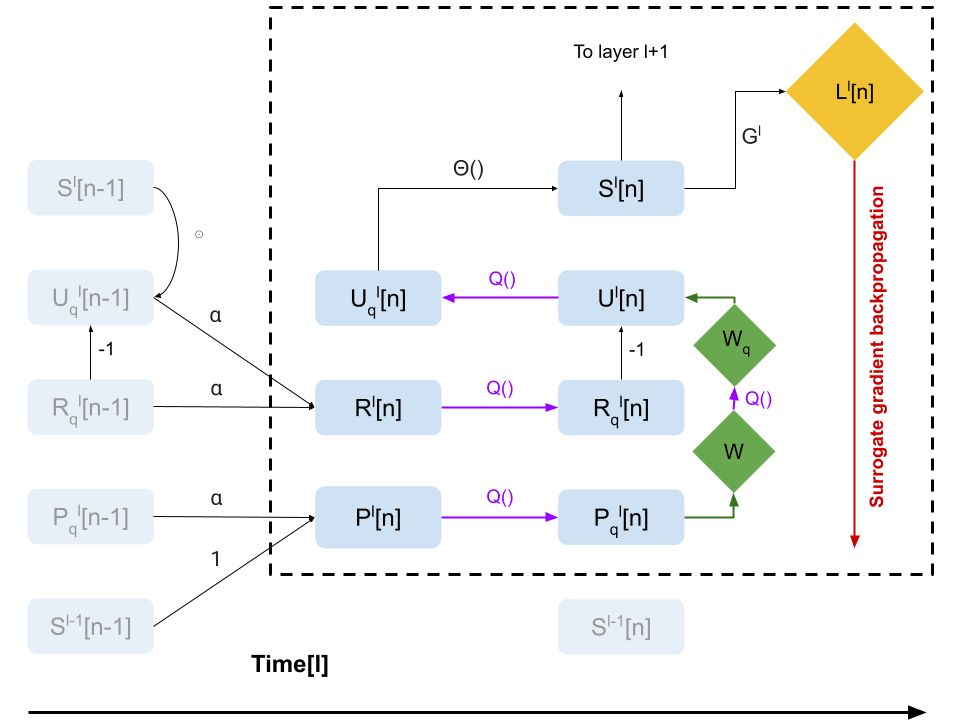}
\caption{State diagram of a quantized \ac{SNN} layer (grouped by dashed box) for one time step. $Q$ quantizes the state variables $P$, $U$, $R$, and the weights $W$.}\label{fig:snn-quantization}
\end{figure}

We used the QPyTorch~\cite{Zhang2019a} library to simulate the effect of low precision fixed-point quantization with various bit-precision settings for both the neuron state variables and synaptic weights. 
Figure~\ref{fig:snn-quantization} shows how the quantization was performed on each layer of the \ac{SNN}.
The quantizer function $Q$ shown in figure~\ref{fig:snn-quantization} used the low precision fixed-point quantizer provided by QPyTorch. 
It performed simulated quantization by rounding or clipping a value to the desired bit-precision, but still used the full precision PyTorch tensor to hold the quantized values. The quantizers used stochastic rounding~\cite{Muller2015RoundingWeights}, with the probability of rounding up or down depending on the distance between the true value to the two nearest quantized values as follows, 
\begin{equation}
\label{eqn:stoachstic_rounding}
\begin{split}
Q(x)
=
\Biggl \lbrace
{
\lfloor x \rfloor, \text{ with probability } 1 - \frac{x - \lfloor x \rfloor}{\epsilon}
\atop
 \lfloor x \rfloor + \epsilon, \text{ with probability } \frac{x - \lfloor x \rfloor}{\epsilon}
},
\end{split}
\end{equation}
where $\epsilon$ was the smallest represented value of a given bit-precision, and $\lfloor x \rfloor$ is the rounded-down value. This allowed the discarded precision to be represented by a stochastic process. We also enabled value clamping, whereby values outside of the represented range were clamped to the maximum or minimum represented value. For this, the backward loss gradient was zeroed if the forward value was clamped. Without gradient zeroing, there would be a mismatch between the loss gradient and the forward clamped value. Finally, due to the reduced precision of the weights, a gradient scaling factor of $1e3$ was applied to magnify the step size of the weight updates. This avoided the updates being much smaller than $\epsilon$ of a given bit-precision.

\section{Hessian trace for layer-wise bit precision quantization} \label{hessian}
The Hessian matrix is the second order differential of the loss with respect to the weights, and represents the sensitivity of the loss to any perturbation of the weights. 
Suppose that a fully converged network is at a global or local minimum of the loss space, the gradient at this point should be close to zero. 
Therefore the gradient itself does not provide information on how weight quantization affects the loss. 
However, if the converged network has a Hessian matrix with very large eigenvalues, any small perturbation of the weights would result in large loss gradients, and hence the loss itself. 
Therefore, the relative magnitude of the Hessian of each layer can tell us how the numerical errors introduced by weight quantization at that layer would lower the accuracy of the network. We can use this information to assign an optimal bit-precision to each layer according to each layer's Hessian.

Except for very small networks, computing the entire Hessian matrix is prohibitive.
However, for the purpose of mixed precision quantization, the Hessian \emph{trace} has been found to be sufficient~\cite{Dong2019}. 
The Hessian trace is the sum of all eigenvalues of the Hessian matrix, and can be evaluated by randomized numerical linear algebra methods. 
In particular, the Hutchinson algorithm~\cite{Avron2011} enables fast estimations of the trace using only the Hessian-vector product. For any symmetric matrix $H \in R^{d \times d}$, and a random vector $v \in R^d$ with i.i.d components sampled from a Gaussian distribution ($\mathcal{N}(0, 1)$), the trace of the matrix can be computed by the expectation of vector-matrix-vector product as follows,
\begin{equation}
\label{eqn:hessian_trace}
\begin{split}
Tr(H) = \mathbb{E}[v^t H v].
\end{split}
\end{equation}
Algorithm~\ref{alg:hutchinson} shows the Hessian trace computation adapted for our \ac{SNN} model.
Note that the Hessian matrix is never computed explicitly, and the PyTorch autograd function is used to evaluate the Hessian-vector product $Hv$ for each batch of the data and for each time step.

\begin{algorithm}[h]
    \caption{Computing the Hessian Trace of an SNN layer
    \label{alg:hutchinson}}
    \begin{algorithmic}[1]
    \REQUIRE {Layer parameters $W_i$}
    \STATE $vHv \gets []$
    \FOR{$i \gets 1 \textrm{ to } MaxIter $}
        \STATE Initialize a random vector $v$ with same shape as $W_i$ 
         \STATE Initialize a zero vector $z$ with same shape as $W_i$ 
         \STATE Normalize $v$, $v=\frac{v}{\lVert v \rVert_2}$
        \FOR{$b \gets 1 \textrm{ to } MaxBatch$}
            \FOR{$n \gets 1 \textrm{ to } MaxSeq$}
                 \STATE Compute $g_i = \frac{\partial \mathcal{L}}{\partial W_i}$ by autograd
                \STATE Compute $Hv = \frac{\partial (g_i^T v)}{\partial W_i}$ by autograd
                \STATE Accumulate $z \gets z + Hv$
            \ENDFOR
        \ENDFOR
        \STATE Append $v^T z$ to $vHv$
    \ENDFOR
    \STATE $trace \gets mean(vhv)$
    \STATE \RETURN $trace$ 
\end{algorithmic}
\end{algorithm}

\section{Experiments and Results} \label{results}
We performed all experiments on the N-MNIST~\cite{Orchard_etal15_convstat} hand written digit recognition dataset that was recorded with an event based camera \cite{Lichtsteiner2008}. We used the same network architecture as the original DECOLLE network~\cite{Kaiser2020}, which consisted of 3 convolutional layers with filter size $7 \times 7$ and 64, 128, 128 channels respectively. Instead of the original neuron model with 5 state variables, the simplified model of (\ref{eqn:srm}) was used. All network configurations and hyper-parameters were kept the same, except for $lr=1e6$ and $\alpha=0.97$ as they were more suitable for the simplified neuron model. The same train-test split of the dataset was used, with 2000 samples in the training set and 100 samples in the testing set for each of the 10 classes. We first trained a network with full precision without any quantization for 20 epochs. The Hessian trace for each layer was then computed using algorithm \ref{alg:hutchinson}. We then fine-tuned the same network with quantization as shown in figure~\ref{fig:snn-quantization} for another 10 epochs, using a layer-wise bit-precision for each layer. The experiments were repeated 3 times with different weight initializations, and the mean and standard deviation of the accuracy were reported.

Table~\ref{tab:trace} shows the Hessian trace of each layer after 20 epochs of full precision training. The Hessian trace increased from L1 to L3, with 4 orders of magnitude difference. However, there was only 1 order of magnitude difference between L2 and L3, indicating relatively equal sensitivity to weight perturbation between these two layers. This increasing trend of Hessian trace is compatible with the trend of the accuracy of each layer's local classifier, which also increased from L1 to L3.

\begin{table}[htbp]
\caption{\label{tab:trace}Hessian Trace and accuracy of different SNN layers }
\centering
\begin{tabular}{ cccc } 
    \toprule
    & L1 & L2 & L3 \\
    \midrule
    Trace & 3.57e2 & 2.86e5 & 3.45e6 \\
    Accuracy & 47.8 & 95.0 & 98.2 \\
  \bottomrule
\end{tabular}
\end{table}

Table~\ref{tab:accuracy} shows the e accuracy of the final layer after fine tuning the network with layer-wise bit-precision quantization, and the corresponding size of the network parameters after the quantization. One surprising observation is that assigning a 16-bit-precision for each layer did not lower the accuracy at all, while reducing the network size by 50\%. This shows that full precision arithmetic is not needed. As expected, the accuracy dropped when lowering the bit-precision of all 3 layers from 16 to 8 and 4 bits, from 98.1\% to 96.3\% and 81.1\% respectively. We also compared the accuracy when using a layer-wise bit-precision, with a (16, 8) and (8, 4) bit combination. As indicated by the Hessian trace shown on table~\ref{tab:trace}, we expected assigning L1 with a low bit-precision to have the least impact to accuracy, followed by L2 and L3. This was confirmed by the results, with the accuracy dropping from 98.0\% to 97.9\% and 97.6\% as we moved an 8-bit quantizer from L1 to L3 in a 16-bit network. The same observation was found when moving a 4-bit quantizer from L1 to L3 in an 8-bit network, with the accuracy dropping from 95.0\% to 94.0\% and 93.6\%. 
Finally we quantized the first two layers of the network with lower bit-precision, and found that the (8, 8, 16) combination provided a good balance between accuracy and network size, with the accuracy dropping by only 0.3\% while the network size was reduced by 58\%.

\begin{table}[htbp]
\caption{\label{tab:accuracy}Accuracy of the final SNN layer after fine tuning }
\centering
\begin{tabular}{ cccccc } 
    \toprule
    L1 & L2 & L3 & Accuracy & Size (MB) \\
    \midrule
    FP32 & FP32 & FP32 & 98.1 $\pm$ 0.2 & 4.84 \\
    \midrule
    16 & 16 & 16 & 98.1 $\pm$ 0.2 & 2.42\\
    \midrule
    8 & 16 & 16 & 98.0 $\pm$ 0.2 & 2.41 \\
    16 & 8 & 16 & 97.9 $\pm$ 0.1 & 2.02 \\
    16 & 16 & 8 & 97.6 $\pm$ 0.2 & 1.62 \\
    \midrule
    8 & 8 & 16 & 97.8 $\pm$ 0.2 & 2.01 \\
    8 & 8 & 8 & 96.3 $\pm$ 0.3 & 1.21 \\
    \midrule
    4 & 8 & 8 & 95.0 $\pm$ 0.5 & 1.21 \\
    8 & 4 & 8 & 94.0 $\pm$ 0.9 & 1.01 \\
    8 & 8 & 4 & 93.6 $\pm$ 0.8 & 0.81 \\
    \midrule
    4 & 4 & 8 & 92.6 $\pm$ 1.1 & 1.01 \\
    4 & 4 & 4 & 81.1 $\pm$ 1.0 & 0.61 \\
  \bottomrule
\end{tabular}
\end{table}
\section{Conclusion and future work} \label{conculsion}
In this work we presented a method of quantizing \acp{SNN} for high throughput, energy efficient inference. We used a simplified LIF neuron model that has two equivalent forms, with one form being compatible with gradient based DECOLLE training, and another form suitable for inference with reduced computation and memory usage. We also applied Hessian aware quantization on SNN and used the layer-wise Hessian trace to evaluate the sensitivity of the loss to the quantization of each layer's weights. We found that the Hessian trace increased 4-fold from layer L1 to L3, and confirmed that this provided good information to determine the optimal layer-wise bit-precision for quantization. Our results show that when assigning a low bit-precision to a layer with low Hessian trace, it had a smaller impact on accuracy than reducing the bit-precision of a layer with a high Hessian trace.

There are a number of limitations of the present work. The network only had 3 layers and could not take full advantage of the Hessian aware quantization method. However, this is the limitation of \acp{SNN} in general, as having many layers has not been shown to result in a higher accuracy using the available benchmarks~\cite{Kaiser2020}. After the Hessian trace analysis, the bit-precision allocation step was performed manually, due to the small number of layers available, as opposed to the Pareto Frontier method used in \cite{Dong2019}. Finally, the present work was performed with simulated quantization provided by QPyTorch, and the underlying computation was still performed in FP32. Future work should explore using Hessian aware quantization of SNNs with native fixed-point arithmetic on a neuromorphic hardware to more accurately assess the trade off between bit-precision, compute time, memory usage and energy efficiency.

\section{Acknowledgements} \label{acks}
This work was supported by the Korean Institute for Science and Technology (EN), the National Science Foundation under grant 1652159 (EN) and 1823366 (HL, EN).

\bibliographystyle{unsrt}
\bibliography{references,biblio_unique_alt}

\begin{thebibliography}{10}

\bibitem{Pfeiffer2018}
Michael Pfeiffer and Thomas Pfeil.
\newblock {Deep Learning With Spiking Neurons: Opportunities and Challenges}.
\newblock {\em Frontiers in Neuroscience}, 2018.

\bibitem{Indiveri2011}
Giacomo Indiveri, Bernabé Linares-Barranco, Tara~Julia Hamilton, André van
  Schaik, Ralph Etienne-Cummings, Tobi Delbruck, Shih~Chii Liu, Piotr Dudek,
  Philipp H{\"{a}}fliger, Sylvie Renaud, Johannes Schemmel, Gert Cauwenberghs,
  John Arthur, Kai Hynna, Fopefolu Folowosele, Sylvain Saighi, Teresa
  Serrano-Gotarredona, Jayawan Wijekoon, Yingxue Wang, and Kwabena Boahen.
\newblock {Neuromorphic silicon neuron circuits}, 2011.

\bibitem{Esser_etal16_convnetw}
Steven~K Esser, Paul~A Merolla, John~V Arthur, Andrew~S Cassidy, Rathinakumar
  Appuswamy, Alexander Andreopoulos, David~J Berg, Jeffrey~L McKinstry, Timothy
  Melano, Davis~R Barch, et~al.
\newblock Convolutional networks for fast, energy-efficient neuromorphic
  computing.
\newblock {\em PNAS}, 113:11441--11446, 2016.

\bibitem{Davies_etal18_loihneur}
M.~Davies, N.~Srinivasa, T.~H. Lin, G.~Chinya, P.~Joshi, A.~Lines, A.~Wild, and
  H.~Wang.
\newblock Loihi: A neuromorphic manycore processor with on-chip learning.
\newblock {\em IEEE Micro}, PP(99):1--1, 2018.

\bibitem{Bartol2015}
Thomas Bartol, Cailey Bromer, Justin Kinney, Michael Chirillo, Jennifer Bourne,
  Kristen Harris, and Terrence Sejnowski.
\newblock {Hippocampal Spine Head Sizes are Highly Precise}.
\newblock {\em bioRxiv}, 2015.

\bibitem{Schaefer2020}
Clemens~J.S. Schaefer and Siddharth Joshi.
\newblock {Quantizing Spiking Neural Networks with Integers}.
\newblock In {\em ACM International Conference Proceeding Series}, 2020.

\bibitem{Esser2019}
Steven~K. Esser, Jeffrey~L. McKinstry, Deepika Bablani, Rathinakumar Appuswamy,
  and Dharmendra~S. Modha.
\newblock {Learned step size quantization}, 2019.

\bibitem{Dong2019}
Zhen Dong, Zhewei Yao, Yaohui Cai, Daiyaan Arfeen, Amir Gholami, Michael~W.
  Mahoney, and Kurt Keutzer.
\newblock {HAWQ-V2: Hessian aware trace-weighted quantization of neural
  networks}, 2019.

\bibitem{Gerstner2014}
Wulfram Gerstner, Werner~M. Kistler, Richard Naud, and Liam Paninski.
\newblock {\em {Neuronal dynamics: From single neurons to networks and models
  of cognition}}.
\newblock 2014.

\bibitem{Zenke2021}
Friedemann Zenke and Emre~O. Neftci.
\newblock {Brain-Inspired Learning on Neuromorphic Substrates}.
\newblock {\em Proceedings of the IEEE}, 2021.

\bibitem{Kaiser2020}
Jacques Kaiser, Hesham Mostafa, and Emre Neftci.
\newblock {Synaptic Plasticity Dynamics for Deep Continuous Local Learning
  (DECOLLE)}.
\newblock {\em Frontiers in Neuroscience}, 2020.

\bibitem{Gerstner_etal14_neurdyna}
Wulfram Gerstner, Werner~M Kistler, Richard Naud, and Liam Paninski.
\newblock {\em Neuronal dynamics: From single neurons to networks and models of
  cognition}.
\newblock Cambridge University Press, 2014.

\bibitem{Zhang2019a}
Tianyi Zhang, Zhiqiu Lin, Guandao Yang, and Christopher de~Sa.
\newblock {QPyTorch: A low-precision arithmetic simulation framework}, 2019.

\bibitem{Muller2015RoundingWeights}
Lorenz~K. Muller and Giacomo Indiveri.
\newblock {Rounding Methods for Neural Networks with Low Resolution Synaptic
  Weights}.
\newblock 4 2015.

\bibitem{Avron2011}
Haim Avron and Sivan Toledo.
\newblock {Randomized algorithms for estimating the trace of an implicit
  symmetric positive semi-definite matrix}.
\newblock {\em Journal of the ACM}, 2011.

\bibitem{Orchard_etal15_convstat}
Garrick Orchard, Ajinkya Jayawant, Gregory~K. Cohen, and Nitish Thakor.
\newblock Converting static image datasets to spiking neuromorphic datasets
  using saccades.
\newblock {\em Frontiers in Neuroscience}, 9, nov 2015.

\bibitem{Lichtsteiner2008}
Patrick Lichtsteiner, Christoph Posch, and Tobi Delbruck.
\newblock {A 128 × 128 120 dB 15 {$\mu$}s latency asynchronous temporal
  contrast vision sensor}.
\newblock {\em IEEE Journal of Solid-State Circuits}, 2008.

\end{thebibliography}

\end{document}